\documentclass{llncs}

\usepackage{graphicx,times,amsmath} 
\usepackage{url}
\usepackage{graphicx}
\usepackage{amsfonts}
\usepackage{wasysym}
\usepackage{latexsym}
\usepackage{color}
\usepackage{multirow}
\usepackage{amssymb}
\usepackage{epstopdf}

\usepackage{array}
\usepackage{multirow} 
\usepackage{verbatim}
\usepackage{amstext}
\usepackage{amsmath}
\usepackage{rotating}
\usepackage{graphicx}
\usepackage{amssymb}
\usepackage{subfigure}
\usepackage[svgnames]{xcolor}
\usepackage{float} 
\usepackage[ruled,vlined]{algorithm2e}
\usepackage{multirow}
\usepackage{url}
\usepackage{colortbl}
\usepackage{caption}

\begin{document}

\title {Computing the Dirichlet-Multinomial Log-Likelihood Function}





\author{xx xx\inst{1}$^{,}$\inst{2} xx xx \inst{2}}

\authorrunning{Allesiardo et al.} 

\institute{xxx,\\
xxx\\
\and xxx}

\author{Djallel Bouneffouf\inst{1}}

\authorrunning{Allesiardo et al.} 

\institute{IBM Research}

\maketitle 

\begin{abstract}
Dirichlet-multinomial (DMN) distribution is commonly used to model over-dispersion in count data. Precise and fast numerical computation of the DMN log-likelihood function is important for performing statistical inference using this distribution, and remains a challenge. To address this, we use mathematical properties of the gamma function to derive a closed form expression for the DMN log-likelihood function. Compared to existing methods, calculation of the closed form has a lower computational complexity, hence is much faster without comprimising computational accuracy. 
\end{abstract}


\section{Introduction}

The uses of the binomial and Multinomial (MN) distributions in statistical modelling are very important, with a huge variety of applications in bioinformatics \cite{Brown93usingdirichlet}. The modeling of over-dispersion of the MN distribution has been solved by extending the MN distribution to the Dirichlet-multinomial (DMN) distribution \cite{Mosimann1962}.

The computation of the DMN log-likelihood function by existing methods leads to non accurate results \cite{yu2014efficient}. Recently, an alternative formulation done in \cite{yu2014efficient} solve this issue, but it still leads to long runtimes that make it inappropriate for large count data. We have developed a new method for computation of the DMN log-likelihood to solve the accuracy problem without incurring long runtimes. 

The new approach is mainly based on the definition of the Dirichlet distribution according to the Gamma function. Our numerical experiments show that this new method improves the runtime in high-count data situations that are common in deep sequencing data, and then increases the usefulness of the DMN distribution to model many bioinformatics problems.

\section{Previous work}

Due to its demonstrated effectiveness and its mathematical convenience, the DMN distribution is widely applied into different domains, such as multiple sequence alignment \cite{Brown93usingdirichlet} in bioinformatics, natural language processing \cite{mimno2012topic} \cite{mackay1995hierarchical}, finance \cite{madsen2005modeling} and online learning \cite{BouneffoufBG13,ChoromanskaCKLR19,RiemerKBF19,LinC0RR20,lin2020online,lin2020unified,NoothigattuBMCM19,surveyDB,LR85,Bouneffouf0SW19,LinBCR18,DB2019,BalakrishnanBMR19ibm,BouneffoufLUFA14,RLbd2018,balakrishnan2020constrained,BouneffoufRCF17,BalakrishnanBMR18,BouneffoufBG12,Bouneffouf16,aaai0G20,AllesiardoFB14,dj2020,Sohini2019,bouneffouf2020online}  . 

Likelihood functions play a important role in statistical, where it is generally used for parameter estimation and interval estimation \cite{casella2002statistical}. 
Unfortunately, numerical computation of the DMN log-likelihood function by conventional methods results in instability in the neighborhood of $\phi=0$, where the parameter $\phi$ characterizes how different a DMN distribution is from the corresponding MN distribution with the same category probabilities, and gives the DMN distribution the ability to capture variation that cannot be done by the MN distribution. 
Hence, the functions implementing of the DMN done in \cite{CoreTeam2013}, \cite{tvedebrink2009dirmult} or \cite{lesnoff2012aod}, return NaN when $\phi=0$.

To overcome this instability of accuracy on computing the log-likelihood, authors in \cite{yu2014efficient} derived a new approximation of the DMN log-likelihood function based on Bernoulli polynomials. Using a mesh algorithm, they computed the log-likelihood for any parameters in the domains of the log-likelihood function. Comparing with the existing methods, the proposed method is more accurate and is faster. 

In this paper we propose a much faster method for the computation of the DMN log-likelihood function and we demonstrate it's efficiency though simulation. 

\section{DMN DISTRIBUTION}
Following the notation done in \cite{yu2014efficient}, the probability mass function (PMF) of the $K$ categories MN distribution of $N$-independent trials is given by

\begin{equation}
 \label{eq:pr20}
f_{MN}(x;N,p)= \frac{N!}{\prod^K_{k=1}x_k!}\prod^K_{k=1}p^{x_k}_k
\end{equation} 

where $N!$ denotes the factorial of a non-negative integer $N$; the observations $x=(x_1,...,x_k)$, satisfying $\sum^K_{k=1} x_k=N $, are non-negative integers; and $p=(p_1,...,p_k)$, satisfying $\sum^K_{k=1} p_k=1 $, are the probabilities that these $K$ categories occur.

The DMN distribution is generated with the probabilities $p$ following a prior distribution (of the positive parameters $\alpha=(\alpha_1,...,\alpha_K) $) conjugate to the PMF $f_{MN}(x;N,p)$ \cite{bishop2007pattern} as follows:

\begin{equation}
 \label{eq:pr21}
f_{Dir}(p;\alpha) \propto \prod^K_{k=1}p^{\alpha_i-1}_k
\end{equation} 

The normalized form of the Dirichlet distribution is as follows: 

\begin{equation}
 \label{eq:pr24}
f_{Dir}(p;\alpha) = \frac{N!}{\prod^K_{k=1}x_k!}\prod^K_{k=1}p^{x_k}_k
\end{equation} 

where $\Gamma(x)$ is the gamma function and $A=\sum^K_{i=1} \alpha_i$.

By taking the integral of the product of the MN likelihood (1) the Dirichlet prior (2) and with respect to the probabilities $p$, the PMF of the DMN distribution is derived \cite{Mosimann1962}, 
\begin{equation}
 \label{eq:pr22}
f_{Dir}(x:N;\alpha) = \frac{N!}{\prod^K_{k=1}x_k!} \frac{\Gamma(A)}{\Gamma(A+N)}\prod^K_{k=1}\frac{\Gamma(\alpha_k+x_k)}{\Gamma(\alpha_k)} 
\end{equation} 

where, same as the MN distribution, $x=(x_1,...,x_K)$ are non-negative integers, satisfying $N=\sum^K_{k=1}k$. 
The first term on the right side of (3) does not depend on the parameter $\alpha$. For common uses of the likelihood function in the maximum-likelihood estimation, we are only interested in the last two terms of the DMN likelihood function in (3), then: 

\begin{equation}
 \label{eq:pr23}
L(\alpha:x) = \frac{\Gamma(A)}{\Gamma(A+N)}\prod^K_{k=1}\frac{\Gamma(\alpha_k+x_k)}{\Gamma(\alpha_k)} 
\end{equation} 

\section{PROPOSITION}

To compute the log-likelihood of the DMN we are using the definition of the Dirichlet-distribution. 
According to the definition of the $\Gamma$ function we have,

\begin{equation}
 \label{eq:pr26}
\Gamma(A)= (A-1)\Gamma(A-1) 
\end{equation} 

then

\begin{equation}
 \label{eq:pr26}
\frac{\Gamma(A)}{\Gamma(A+N)}= (\prod_{i=A}^{A+N-1} i)^{-1}  
\end{equation} 

and

\begin{equation}
 \label{eq:pr261}
 \frac{\Gamma(\alpha_k + x_k)}{\Gamma(\alpha_k)} =\prod_{i=\alpha_k}^{\alpha_k+x_k-1}i
\end{equation} 

then

\begin{equation}
 \label{eq:pr27}
 L(\alpha,x)=\frac{\prod_{k=1}^{K} \prod_{j=\alpha_k}^{\alpha_k + x_k}j}{\prod_{k=A}^{A+N-1} i} 
\end{equation}

By taking the logarithm of both sizes of (\ref{eq:pr27}), we get the log-likelihood
function
\begin{equation}
 \label{eq:pr28}
\log(L(\alpha,x))= \sum_{k=1}^{K} \sum_{j=\alpha_k}^{\alpha_k + x_k-1} log(j) - \sum_{k=A}^{A+N-1} log(i) 
\end{equation}
We call this new method of computing the Dirichlet-multinomial distribution "ExactDMN".

\subsection{Complexity}

Table \ref{table:Synthetic} gives the complexity of each methods. 

\begin{table}[ht]
\centering
\caption{The DMN method's complexity}
\begin{tabular}{|l|c| }
  \hline
  Methods & Complexity  \\
  \hline
  VGAM \cite{yee2010vgam} & $O(A+N-1)+O(A-1)+$  \\ 
        & $\sum^K_{k=1}O(\alpha_k+x_k+\alpha_{k-2})$  \\ 
  Mesh \cite{yu2014efficient} & $O(m)+O(\sum^K_{k=1}(m+x_k))$  \\
  ExactDMN & $O(\sum^K_{k=1}(x_k-1))+O(N-1)$   \\

  \hline
\end{tabular}
    \label{table:Synthetic} 
 \end{table}

We observe from the the table that $ExactDMN$ implementation is less time consuming then the $VGAM$ and Mesh implementations, with $m >N$ and $m$  .

\section{Experiments}

We implemented the algorithm for computing the DMN log-likelihood in java. In this section, we demonstrate the accuracy and run-time of the proposed algorithm. All experiments were run on a Linux machine with a 4-core Intel Xeon CPUs E5630@ 3.53GHz. 

We compute the error of the ExactDMN algorithm by comparing its results with the results of an implementation of (4) in \cite{steinsage}, which can achieve high accuracy \cite{yu2014efficient}.

\begin{figure}[ht]
  \caption{DMN Method's error with log-lik error ($10^{-14}$)}
   \label{fig:error} 
  \centering
    \includegraphics[width=1.09\textwidth]{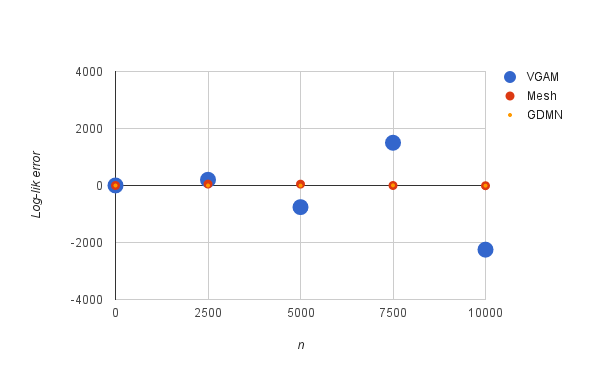}
\end{figure}

Figure~\ref{fig:error} compares the error of the DMN methods where the parameters are
$x=n(1,1,1,1)$, $\phi=1/200$ and $p=(.1,.2,.3,.4)$. We can see the ExactDMN algorithm is more accurate then VGAM and has the same accuracy as mesh algorithm.

\begin{figure}[ht]
  \caption{DMN Method's runtime}
   \label{fig:runtime} 
  \centering
    \includegraphics[width=1.09\textwidth]{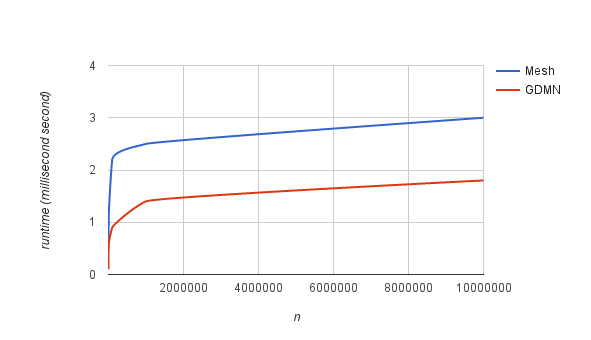}
\end{figure}

Figure~\ref{fig:runtime} shows that the runtime of the DMN methods with the following parameters, $x=n(1,2,3)$, $p=(1/6,1/3,1/2)$ and $\phi=1/60$, where each plot represents $100$ DMN log-likelihood evaluations.
We can see from the Figure~\ref{fig:runtime} that the ExactDMN algorithm is much faster than the mesh algorithm for the DMN log-likelihood computation. DMNG algorithm increases more slowly as the counts $n$ increase than the method in mesh. 

\section{Conclusion}
Over-dispersion is important and needs to be accommodated in modeling count data. To handle over-dispersion in MN data, the DMN distribution is commonly used, but its computation is still a challenge.
Mainly based on the definition of the Dirichlet distribution, our method solves the runtime challenge without undergoing errors. 

\vspace{-2mm}
\bibliographystyle{splncs}
\bibliography{biblio}
\end{document}